\documentclass[conference]{IEEEtran}
\IEEEoverridecommandlockouts
\usepackage{cite}
\usepackage{amsmath,amssymb,amsfonts}
\usepackage{algorithmic}
\usepackage{graphicx}
\usepackage{textcomp}
\usepackage{booktabs}
\usepackage{multirow,censor, mathrsfs, dsfont}
\usepackage{xcolor}
\usepackage{soul}

\def\BibTeX{{\rm B\kern-.05em{\sc i\kern-.025em b}\kern-.08em
    T\kern-.1667em\lower.7ex\hbox{E}\kern-.125emX}}
\begin{document}
\title{Development of a face mask detection pipeline for mask-wearing monitoring in the era of the COVID-19 pandemic: A modular approach}

\author{\IEEEauthorblockN{Benjaphan Sommana\IEEEauthorrefmark{1}, Ukrit Watchareeruetai,\\Ankush Ganguly, Samuel W.F. Earp}
\IEEEauthorblockA{Sertis Vision Lab \\597/5 Sukhumvit Road, Watthana, \\Bangkok 10110 Thailand\\
Email: \IEEEauthorrefmark{1}bsomm@sertiscorp.com}
\and
\IEEEauthorblockN{Taya Kitiyakara, Suparee Boonmanunt,\\Ratchainant Thammasudjarit}
\IEEEauthorblockA{Faculty of Medicine\\ Ramathibodi Hospital, Mahidol University\\
270 Rama VI road, Thung Phaya Thai, Ratchathewi\\ Bangkok 10400 Thailand}}

\IEEEoverridecommandlockouts
\IEEEpubid{\makebox[\columnwidth]{978-1-6654-3831-5/22/\$31.00~\copyright2022 IEEE \hfill} \hspace{\columnsep}\makebox[\columnwidth]{ }}

\maketitle

\begin{abstract}
During the SARS-Cov-2 pandemic, mask-wearing became an effective tool to prevent spreading and contracting the virus. The ability to monitor the mask-wearing rate in the population would be useful for determining public health strategies against the virus.
In this paper, we present a two-step face mask detection approach consisting of two separate modules: 1) face detection and alignment and
2) face mask classification.
This approach allows us to experiment with different combinations of face detection and face mask classification modules.
More specifically, we experimented with PyramidKey and RetinaFace as face detectors while maintaining a lightweight backbone for the face mask classification module. 
Moreover, we also provide a relabeled annotation of the test set of the AIZOO dataset, where we rectified the incorrect labels for some face images.
The evaluation results on the AIZOO and Moxa 3K datasets show that the proposed face mask detection pipeline surpassed the state-of-the-art methods.
The proposed pipeline also yielded a higher mAP on the relabeled test set of the AIZOO dataset than the original test set.
Since we trained the proposed model using in-the-wild face images, we can successfully deploy our model to monitor the mask-wearing rate using public CCTV images.
\end{abstract}


\IEEEpeerreviewmaketitle

\section{Introduction} \label{sec:intro}
Since the beginning of the SARS-Cov-2 pandemic in the later part of 2019, there have been more than five million deaths from the disease globally and many more hospitalized\footnote{https://covid19.who.int/}.
Many countries have suffered economic problems due to repeated lockdowns and decreased consumption. 
As the SARS-Cov-2 virus is a respiratory virus and spreads by a mixture of large droplets and airborne particles \cite{liu_2020, jayaweera_2020}, non-pharmaceutical interventions such as mask-wearing, hand-washing, and social distancing have been used to limit the spread of the virus\footnote{https://www.who.int/emergencies/diseases/novel-coronavirus-2019/advice-for-public}.
A recent meta-analysis has suggested that wearing masks could reduce the spread of the virus by up to 50\% \cite{Talic_2021}. 
Most studies evaluating the rate of mask-wearing have used the presence of mask mandates or online questionnaires \cite{van_2020, lyu_wehby_2020}, or used one-off documentation of mask-wearing data \cite{elachola_2020}.
These have their limitations in terms of accuracy and monitoring the changes in the mask-wearing rate. 
Many published reports have shown that artificial intelligence (AI) and image analysis technologies are capable of recognizing and differentiating faces with and without masks (e.g., \cite{aizootech, Jiang2020, Joshi_2020, roy2020, prasad2021, fan2021}).

\begin{figure*}
    \centering
    \includegraphics[width=0.7\textwidth]{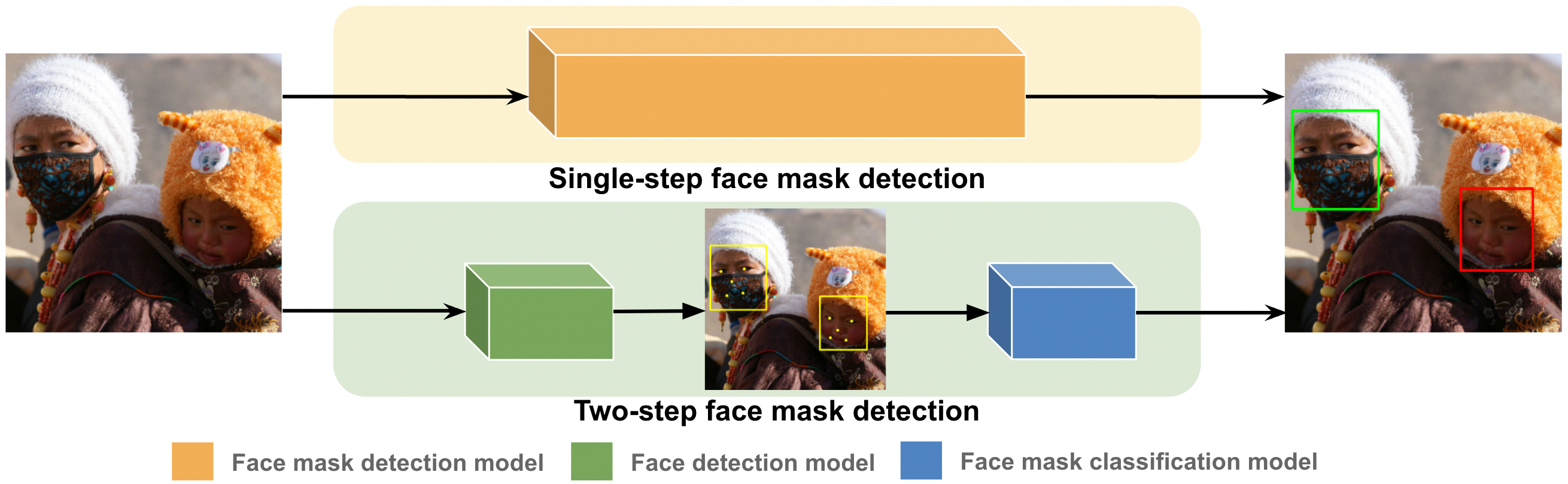}
    \caption{The comparison between single-step and two-step face mask detection pipelines.}
    \label{fig:detection_pipeline}
\end{figure*}

Since 2020, many researchers have explored and developed different face mask detection methods that can be categorized into two main approaches, i.e., single-step and two-step face mask detection.
Recently, single-step face mask detection has gained popularity in the research field with its state-of-the-art performance \cite{aizootech, Jiang2020, roy2020, prasad2021, fan2021}.
This approach trains a single model to localize faces in an input image and determine whether the faces are wearing masks at the same time, as shown in Figure \ref{fig:detection_pipeline}.
It performs similarly to general object detection models but focuses on only two classes, i.e., faces with masks and faces without masks.
The advantage of this approach is that it can be trained end-to-end with a training dataset that includes both the bounding boxes and the face mask labels.
However, available face mask datasets, such as AIZOO \cite{aizootech} and Moxa 3K \cite{roy2020}, are provided with limited sizes of only around 8,000 and 3,000 images, respectively.
Fine-tuning the models on these small face mask datasets may affect their face detection capability.

Another approach is two-step face mask detection which exploits two separate models, i.e., face detection and face mask classification.
A two-step approach firstly localizes all faces in the input image using a face detection model and then determines whether each detected face is wearing a mask with a face mask classification model, as shown in Figure \ref{fig:detection_pipeline}.
Since this approach has two separate models trained on their specific datasets, these isolated models allow us to leverage existing large-scale face detection datasets such as WIDER Face for training face detection models.
With this advantage, the face detection models have better face detection capability.
Additionally, another benefit of this approach is its modularity, which makes it easy to change the face detection and mask classification modules.
In this paper, we developed a face mask detection pipeline based on this two-step face mask detection approach.
Firstly, we adopted the state-of-the-art face detection models from Deng \textit{et al.} \cite{deng2019} and Earp \textit{et al.} \cite{earp2019} to obtain face bounding boxes and landmarks.
Then, the detected faces were cropped and aligned according to the bounding boxes and landmarks.
Finally, we predict the mask class from the cropped and aligned face images by a face mask classification model trained on our collected dataset, namely SertisFaceMask, which contains a wide variety of mask types with a balance of mask classes.
We summarize the key contributions of this research as follows: 
\begin{itemize}

\item We propose a two-step face mask detection pipeline using existing state-of-the-art face detection models and a lightweight face mask classification model trained on a dataset containing a wide variety of mask types with a balance of mask classes.

\item We also provide the relabeled annotation of the test set of the AIZOO dataset \cite{aizootech} that we manually relabeled to rectify the incorrect labels for particular face images.

\item We investigate several techniques for training the face mask classification model, such as label smoothing, face alignment, and focusing on the mask region. 

\item We demonstrate that our two-step face mask detection outperformed state-of-the-art approaches on two benchmarking datasets, i.e., AIZOO dataset \cite{aizootech} and on Moxa 3K dataset \cite{roy2020}.
\end{itemize}
\begin{figure}
    \centering
    \includegraphics[width=0.4\textwidth]{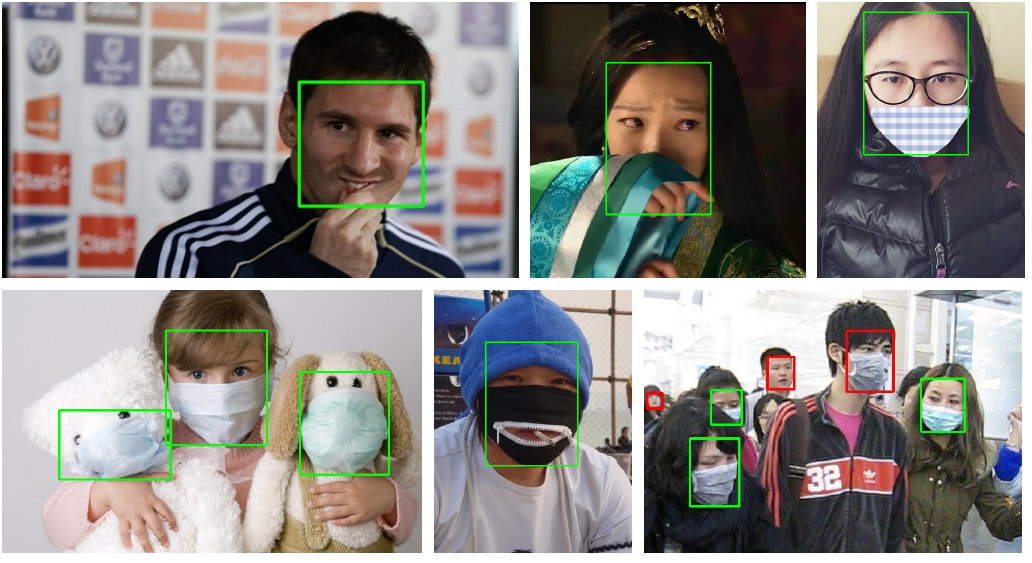}
    \caption{Sample images of incorrect mask class where the green bounding box means face wearing a mask and the red bounding box means faces without masks on a test set of AIZOO dataset.}
    \label{fig:relabel}
\end{figure}

\section{Related work} \label{sec:related}

As mentioned in Section \ref{sec:intro}, we categorize face mask detection into two main approaches, which are single-step and two-step face mask detection.
Most research (e.g., \cite{aizootech, Jiang2020, roy2020, prasad2021, fan2021}) developed face mask detection based on a single-step face mask detection approach.
AIZOO \cite{aizootech} constructed a lightweight face mask detection model exploiting the structure of a Single Shot Detector (SSD) \cite{LiuAESR15} with the backbone of eight convolution layers and trained it on the AIZOO dataset.
Jiang \textit{et al.} \cite{Jiang2020} proposed a face mask detection model, namely RetinaFaceMask, based on RetinaFace \cite{deng2019} structure with ResNet and MobileNet as a backbone.
They exploited the context detection module from Single-Stage Headless (SSH)
\cite{Najibi2018} and applied a convolutional block attention module (CBAM) \cite{Sanghyun2018} to learn discriminated features associated with the face mask.
They used a pre-trained model from WIDER Face and fine-tuned it on the AIZOO dataset.
Roy \textit{et al.} \cite{roy2020} trained the face mask detection models based on popular object detection architectures, e.g., Faster R-CNN \cite{Shaoqing2015}, SSD \cite{LiuAESR15}, YOLOv3, and YOLOv3-tiny \cite{Joseph2018}, pre-trained on the ImageNet dataset \cite{imagenet09}.
Prasad \textit{et al.} \cite{prasad2021} proposed a maskedFaceNet, based on SSD architecture with an RFB network \cite{Liu2017}, with a progressive semi-supervised learning method to handle inadequate training samples.
Fan \textit{et al.} \cite{fan2021} proposed a single-shot lightweight face mask detector with residual context attention module (RCAM) and synthesized Gaussian heatmap regression (SGHR) to gain more discrimination ability.
These single-step methods require a face mask detection dataset to train their models in an end-to-end manner. However, available face mask detection datasets are small in size. For example, AIZOO and Moxa 3K datasets contain only around 8,000 and 3,000 images, respectively. In this work, since we adopt a two-step approach, we can leverage large-scale face detection datasets to train our models.

Very few methods proposed so far adopt a two-step face mask detection pipeline.
In 2020, Joshi \textit{et al.} \cite{Joshi_2020} developed a face mask detection model based on a two-step approach by employing Multi-Task Cascaded Convolutional Networks (MTCNN) \cite{Zhang_2016} for face detection and MobileNetV2 \cite{Sandler2018} for face mask classification.
However, MTCNN is no longer comparable with recent state-of-the-art face detectors.
In this paper, we developed our face mask detection pipeline based on a two-step face mask detection approach because of lacking a large-scale face mask detection dataset.
We adopt state-of-the-art face detection models, RetinaFace \cite{deng2019} and PyramidKey \cite{earp2019}, which yield much better performance on WIDER Face \cite{yang2016} than MTCNN \cite{Zhang_2016}.
Recent state-of-the-art face detectors also predict face landmarks.
In this work, we leverage the predicted landmarks to align the detected faces before feeding into the face mask classification to improve accuracy of classification model.
Besides upgrading the face detection module, we also train the face mask classification model on the collected face mask dataset containing a wide variety of mask types with balance of mask classes and perform the experiment with several training techniques, i.e., label smoothing, face alignment, and focusing on mask region.
%
\section{Datasets} \label{sec:datasets}
This section describes the datasets we used to train and evaluate our face mask classification model and benchmark the mask detection pipeline.
We created an internal dataset named SertisFaceMask to train and evaluate our face mask classification model.
We compared our pipeline with state-of-the-art methods on benchmarking datasets, i.e., AIZOO and Moxa 3K datasets.
\subsection{SertisFaceMask dataset}\label{subsec:sertis_face_mask_dataset}
SertisFaceMask dataset consisted of different mask types, e.g., medical, pollution, gas, and dust masks.
We separated the dataset with more than 30,000 face images into three sets:  training, validation, and test sets with 50\%, 30\%, and 20\%, respectively. 
Each set balanced the number of face images with and without masks.
We collected this dataset from online sources and combined it with occluded face images from Caltech Occluded Faces in the Wild (COFW) dataset \cite{Yue2017}. 
We used PyramidKey \cite{earp2019} to detect faces and then subsequently cropped, aligned and resized face images to 112$\times$112 pixels.

Since we collected this dataset from online sources, it contained some noisy images, e.g., incorrect class, animation, and obscure images, which we manually cleaned from the dataset.
We annotated the dataset by using the visibility of nose as a criterion. When masks covered the nose, images were labeled as `Mask' while masks were below by the nose, images were labeled as `No-Mask'.

\subsection{AIZOO dataset}\label{subsec:face_mask_dataset}
AIZOO \cite{aizootech} provided the face mask detection dataset, called AIZOO dataset, that combines parts of WIDER Face \cite{yang2016} and Masked Faces (MAFA) datasets \cite{ge2017detecting}.
This dataset had two sets: training and test sets. 
The training set contained 6,120 images with 13,593 faces, with almost 78\% of exposed faces.
The test set had 1,830 images with 5,082 faces, with 65\% of faces are not wearing masks.
In order to classify masks that could reduce the spread of the virus, we considered only medical or pollution masks on human faces.
We decided to relabel some faces that were assigned to the incorrect class or wore neither medical nor pollution masks.
We manually corrected the mask class of faces as shown in Figure \ref{fig:relabel}.
We found that we needed to relabel 89 face images.
Among them, we changed the labels of 82 faces from `Mask' to `No-mask', switched five to `Mask', and eliminated two images due to non-human faces.
Moreover, we also released the relabeled annotation of the test set of the AIZOO dataset, which is available online\footnote{https://bit.ly/3E5X1NJ}.
%
\subsection{Moxa 3K dataset}\label{subsec:moxa_3k_dataset}
The Moxa 3K dataset, created by Roy \textit{et al.} \cite{roy2020}, was used for training and evaluating face mask detection models.
This dataset combined images from two different sources; the medical mask dataset from Kaggle and online sources during the COVID-19 pandemic.
This dataset consists of 3,000 images with different scenarios, from close-up faces to crowded scenes, using 2,800 images for training and 200 images for testing.
Unlike the AIZOO dataset, the Moxa 3K dataset contained faces with masks more than exposed faces.
In the training set, only 26\% of the faces were faces without masks, similar to the test set in which about 12\% of faces were without masks.
\section{Proposed method} \label{sec:proposed}
As mentioned in Section \ref{sec:intro}, our pipeline consists of two main modules: 1) face detection and alignment and 2) face mask classification.
As shown in Figure \ref{fig:detection_pipeline}, our pipeline firstly localizes all faces in the input image using a face detection model and then
determines whether each detected face is wearing a mask with a face mask classification model.
We employ the state-of-the-art face detection model from RetinaFace \cite{deng2019} and PyramidKey \cite{earp2019} and train the face mask classification model on our collected dataset.
\subsection{Face detection and alignment}\label{subsec:detection}
As discussed in Section \ref{sec:related}, we exploit several existing face detectors from RetinaFace \cite{deng2019} and PyramidKey \cite{earp2019} which achieved good performance on the validation set of WIDER Face dataset \cite{yang2016}.
RetinaFace and PyramidKey have similar architecture consisting of three parts; visual backbone, neck, and detection head.
The RetinaFace and PyramidKey necks include FPN and independent context modules on the top five and top six levels of pyramid features, respectively.
RetinaFace employs a multi-task loss function that combines face classification, bounding box regression, facial landmark regression, and dense regression loss while PyramidKey considers only the first three losses.
Both face detectors were trained on the WIDER Face dataset \cite{yang2016}.

This paper experiments with several face detection models from RetinaFace using ResNet50 as a visual backbone and PyramidKey using MobileNetV2$_{0.25}$, MobileNetV2$_{1.0}$, and ResNet152 as visual backbones where the model with MobileNetV2$_{0.25}$ backbone of PyramidKey uses only the top three layers of pyramid features while the rest used the top six pyramid features layers.
Additionally, we use a set of five facial landmarks, i.e., eye centers, nose, and mouth corners, obtained from the face detector, to crop and align the detected faces as a pre-processing step before passing the face images to the face mask classification model.


\subsection{Face mask classification}\label{subsec:classification}
To predict the existence of masks on the detected faces, we construct the face mask classification model by using a MobileNetV1$_{1.0}$ \cite{Howard2017} as a backbone. 
The model is pre-trained on the ImageNet dataset \cite{imagenet09} and fine-tuned on the SertisFaceMask dataset.
We adopt Binary Cross Entropy loss which is defined as:
\begin{equation}
    \label{eq:class_loss}
     L = -(y_i log(\hat{y}_i) + (1 - y_i) log(1 - \hat{y}_i)),
\end{equation}
where $y_i$ and $\hat{y}_i$ are ground-truth and softmax probability of $i$-th class ($i \in \{0, 1\}$), respectively.

\section{Experiments}\label{sec:exp}

This section describes different configurations for our face mask detection pipeline and training techniques for the face mask classification model.
We evaluate the proposed method on the test sets of the AIZOO and the Moxa 3K datasets to compare with state-of-the-art approaches.


\subsection{Experimental setup} \label{subsec:setup}
We train the face mask classification model on the training set of the SertisFaceMask dataset by using Stochastic Gradient Descent (SGD) optimizer with the momentum of $0.9$, weight decay of $5 \times 10^{-4}$, and a base learning rate of $10^{-2}$ for 80 epochs reducing it with a factor of $0.1$ at 20, 40, and 60 epochs.
The input images are resized to 224$\times$224 pixels with several augmentation techniques, e.g., cropping with random size and ratio; randomly jittering the brightness, contrast, and saturation; adding AlexNet-style PCA-based noise \cite{krizhevsky_2017}; and horizontal flipping with a probability of $0.5$.
We also apply the dropout technique \cite{srivastava14a} with a dropout rate of $0.5$.
We set a batch size of $32$.
During inference, we detect faces with three detection scales at 320, 640, and 960 and set the face detection threshold and intersection-over-union (IOU) for Non-Maximum Suppression (NMS) to $0.8$ and and $0.5$, respectively.
For face mask classification, we set the input image size to 224 $\times$ 224. 
The predicted class is classified as `Mask' when the probability of the mask existence was equal to or higher than the classification threshold set to $0.95$. 


\subsection{Training techniques}\label{subsec:techniques}

This section investigates training the face mask classification model with several techniques: label smoothing, aligning face with only eye key points, and replacing pixel-level information on the upper half face with random noise or pixels with value zero. 

\subsubsection{Label smoothing} \label{subsec:label_smoothing}
Label smoothing proposed by Szegedy \textit{et al.} \cite{Szegedy15} aims to regularize the classification layer.
The main idea of this technique is to lower the confidence of hard or bad examples for preventing over-fitting and making the model more adaptable.
We manually label hard examples based on face mask visibility to experiment with label smoothing.
We conduct the experiment with two different smoothing factor values of 0.1 and 0.4.

\subsubsection{Aligning with only eye key points} \label{subsec:align_eyes}
The occlusion caused by a face mask affects the localization ability on the nose and mouth corners which may, in turn, affect the precision of face alignment.
Therefore, we also experiment with a modification in the alignment process by aligning with only two key points of the eye centers, which are not occluded while wearing a mask, instead of all five key points.

\subsubsection{Replacing the upper half face} \label{subsec:replace_upper}
The visual appearance of the face mask may be more crucial than the exposed face region for determining whether a mask covers a face in an image.
Consequently, we attempt to test this hypothesis by training a model to focus only on the mask region.
We extract the mask region by leveraging the five key points, i.e., eye centers, nose, and mouth corners, from the face detector. 
In particular, we replace the region above the nose with random noise or zero-valued pixels to ignore the upper half face information and concentrate on the mask appearance only.

\subsection{Evaluation metrics}\label{subsec:metrics}
To compare with other research, we evaluate our face mask detection with precision, recall, F1-score, and mean average precision (mAP).
In order to obtain the mAP, the average precision (AP) of each class needs to be measured.
Specifically, we calculate AP with an all-point interpolation approach \cite{pascal-voc-2012}.

\subsection{Main results}\label{subsec:results}
\subsubsection{Results on the SertisFaceMask dataset} \label{subsec:sertis_result}
Table \ref{tab:techniques} presents the face mask classification results on the test set of the SertisFaceMask dataset with different training techniques.
The incorporation of label smoothing produces a lower F1-score than the baseline setting.
This result shows that applying label smoothing decrease the classification ability and may introduce more ambiguity between the mask and no-mask classes.
Considering face alignment technique, the F1-score for the prediction of no-mask and mask decreases from the baseline setting by about 1.04 and 1.12 points, respectively, when we used face alignment with only eye centers.
Since the images have both exposed and occluded faces, aligning with all key points provides better face images for discrimination than only eye key points.
While the model performs slightly better when the region over the nose was replaced with random noise compared to zero-valued pixels, both of these replacements perform worse than the baseline setting.
This shows that upper face information is also associated with discrimination ability.

\begin{table}[htbp]
\caption{Comparison on F1-score of each class with different techniques.}
\centering
\begin{tabular}{lcc}
\toprule
Technique                                                                            & No-mask        & Mask        \\
\midrule
Baseline                                                                             & \textbf{98.62} & \textbf{98.79} \\
\begin{tabular}[c]{@{}l@{}}Label smoothing, $\epsilon = 0.1$\end{tabular}               & 96.89          & 97.04          \\
\begin{tabular}[c]{@{}l@{}}Label smoothing, $\epsilon = 0.4$\end{tabular}                & 96.79          & 97.16          \\
\begin{tabular}[c]{@{}l@{}}Aligning face with only eye key points\end{tabular}     & 97.58          & 97.67          \\
\begin{tabular}[c]{@{}l@{}}Replacing the upper half face with random noise\end{tabular} & 97.95          & 98.11          \\
\begin{tabular}[c]{@{}l@{}}Replacing the upper half face with zero-valued pixels\end{tabular}  & 97.79          & 97.79         \\
\bottomrule
\end{tabular}
\par
\label{tab:techniques}
\end{table}


\subsubsection{Results on AIZOO dataset} \label{subsec:face_mask_result}
In this section, we compare the performance of the proposed face mask detection pipeline with baseline results from AIZOO \cite{aizootech}, state-of-the-art single-step approaches from RetinaFaceMask \cite{Jiang2020} and SL-FMDet \cite{fan2021}, and two-step approach, MTCNN with  MobileNetV2 \cite{Joshi_2020}.
As shown in Table \ref{tab:face_mask_result}, our two-step face mask detection pipeline gains a large margin of precision on both the mask and no-mask classes compared to the baseline result (5.96--7.79 points).
Compared to RetinaFaceMask, our pipeline with RetinaFace achieves a higher precision in both classes (3.66--17.39 points).
This result can verify that having a separate module between face detection and face mask classification helps to classify the mask class more correctly than using a single model.
Moreover, our pipeline with the smallest PyramidKey - MobileNetV2$_{0.25}$ backbone surpasses RetinaFaceMask with a larger backbone, ResNet50, by a large margin on precision.
Compared to another state-of-the-art single-step approach, SL-FMDet, our pipeline has comparable results with a simpler and more modular pipeline.
In addition, our pipeline also outperforms existing two-step approach, MTCNN with MobileNetV2 \cite{Joshi_2020}.
It shows that exploiting state-of-the-art face detectors improves the ability of face detection by having a higher recall and using efficient face mask classification elevates mask classification ability by gaining higher precision.

We further evaluate our pipeline on the relabeled test set.
The result shows that our pipeline achieves a higher recall of the mask class than the original test set.
This result verifies that our pipeline can detect people with masks effectively and the wrong label of the original test set drops recall in the mask class.



\begin{table*}[htbp]
\caption{The comparative result with the state-of-the-arts on the original and relabeled test set of the AIZOO dataset.}
\centering
\begin{tabular}{llccccccc}
\toprule
\multicolumn{2}{c}{\multirow{2}{*}{Method}}                                                                                                                & \multicolumn{3}{c}{No-mask}                      & \multicolumn{3}{c}{Mask}                         & \multirow{2}{*}{mAP} \\
\cmidrule(l){3-8}
\multicolumn{2}{c}{}                                                                                                                                       & Precision      & Recall         & F1-score       & Precision      & Recall         & F1-score       &                      \\

\midrule
\multicolumn{2}{l}{Baseline \cite{aizootech}}                                                                                                                               & 89.60          & 85.30          & 87.40          & 91.90          & 88.60          & 90.22          & -                    \\
\multicolumn{2}{l}{\begin{tabular}[c]{@{}l@{}}RetinaFaceMask - MobileNet \cite{Jiang2020}\end{tabular}}                                                                                                             & 83.00          & 95.60          & 88.86          & 82.30          & 89.10          & 85.57          & -                    \\
\multicolumn{2}{l}{\begin{tabular}[c]{@{}l@{}}RetinaFaceMask - ResNet \cite{Jiang2020}\end{tabular}}                                                                                                                & 91.90          & \textbf{96.30} & 94.05          & 93.40          & \textbf{94.50} & 93.95          & -                    \\
\multicolumn{2}{l}{\begin{tabular}[c]{@{}l@{}}MTCNN with MobileNetV2 \cite{Joshi_2020}\end{tabular}}                                                                                                                    & 94.50          & 86.38          & 90.26          & 84.39          & 80.92          & 82.62          & -                    \\
\multicolumn{2}{l}{SL-FMDet \cite{fan2021}}                                                                                                                               & -              & -              & -              & -              & -              & -              & \textbf{93.80}       \\
\midrule
\multicolumn{9}{c}{\textbf{Results of our proposed pipeline on the original test set}}                                                                                                                                                                                                  \\
\midrule
\multirow{2}{*}{Face detector}                                           & \multirow{2}{*}{\begin{tabular}[c]{@{}l@{}}Face mask classifier\end{tabular}} & \multicolumn{3}{c}{No-mask}                      & \multicolumn{3}{c}{Mask}                         & \multirow{2}{*}{mAP} \\
\cmidrule(l){3-8}

                                                                         &                                                                                 & Precision      & Recall         & F1-score       & Precision      & Recall         & F1-score       &                      \\
\midrule
\begin{tabular}[c]{@{}l@{}}PyramidKey - MobileNetV2$_{0.25}$  \cite{earp2019}\end{tabular} & \multirow{4}{*}{MobileNetV1$_{1.0}$}    & \textbf{95.56} & 92.73          & 94.12          & 98.84          & 90.01          & 94.22          & 91.09                \\
\begin{tabular}[c]{@{}l@{}}PyramidKey - MobileNetV2$_{1.0}$ \cite{earp2019}\end{tabular}  &                                                                                 & 94.79          & 94.61          & 94.70          & \textbf{99.69} & 91.45          & 95.39          & 92.84                \\
\begin{tabular}[c]{@{}l@{}}RetinaFace - ResNet50 \cite{deng2019}\end{tabular}          &                                                                                 & 95.14          & 95.00          & \textbf{95.07} & 99.59          & 92.60          & \textbf{95.97} & 93.43                \\
\begin{tabular}[c]{@{}l@{}}PyramidKey - ResNet152 \cite{earp2019}\end{tabular}         &                                                                                 & 94.20          & 95.60          & 94.20          & 98.77          & 92.41          & 95.48          & 93.50                \\
\midrule
\multicolumn{9}{c}{\textbf{Results of our proposed pipeline on the relabeled test set}}                                                                                                                                                                                                 \\
\midrule
\begin{tabular}[c]{@{}l@{}}PyramidKey - MobileNetV2$_{0.25}$ \cite{earp2019}\end{tabular} & \multirow{4}{*}{MobileNetV1$_{1.0}$}    & \textbf{95.65} & 91.13          & 93.34          & 98.79          & 93.66          & 96.16          & 92.09                \\
\begin{tabular}[c]{@{}l@{}}PyramidKey - MobileNetV2$_{1.0}$ \cite{earp2019}\end{tabular}  &                                                                                 & 94.90          & 93.09          & 93.98          & \textbf{99.67} & 95.32          & 97.45          & 94.02                \\
\begin{tabular}[c]{@{}l@{}}RetinaFace - ResNet50 \cite{deng2019}\end{tabular}          &                                                                                 & 95.23          & 93.33          & \textbf{94.27} & 99.57          & \textbf{96.36} & \textbf{97.94} & 94.47                \\
\begin{tabular}[c]{@{}l@{}}PyramidKey - ResNet152 \cite{earp2019}\end{tabular}         &                                                                                 & 92.97          & \textbf{93.95} & 93.46          & 98.72          & 96.15          & 97.42          & \textbf{94.58}      \\
\bottomrule
\end{tabular}
\par
\label{tab:face_mask_result}
\end{table*}

\begin{figure}[htbp]
    \centering
\includegraphics[width=0.4\textwidth]{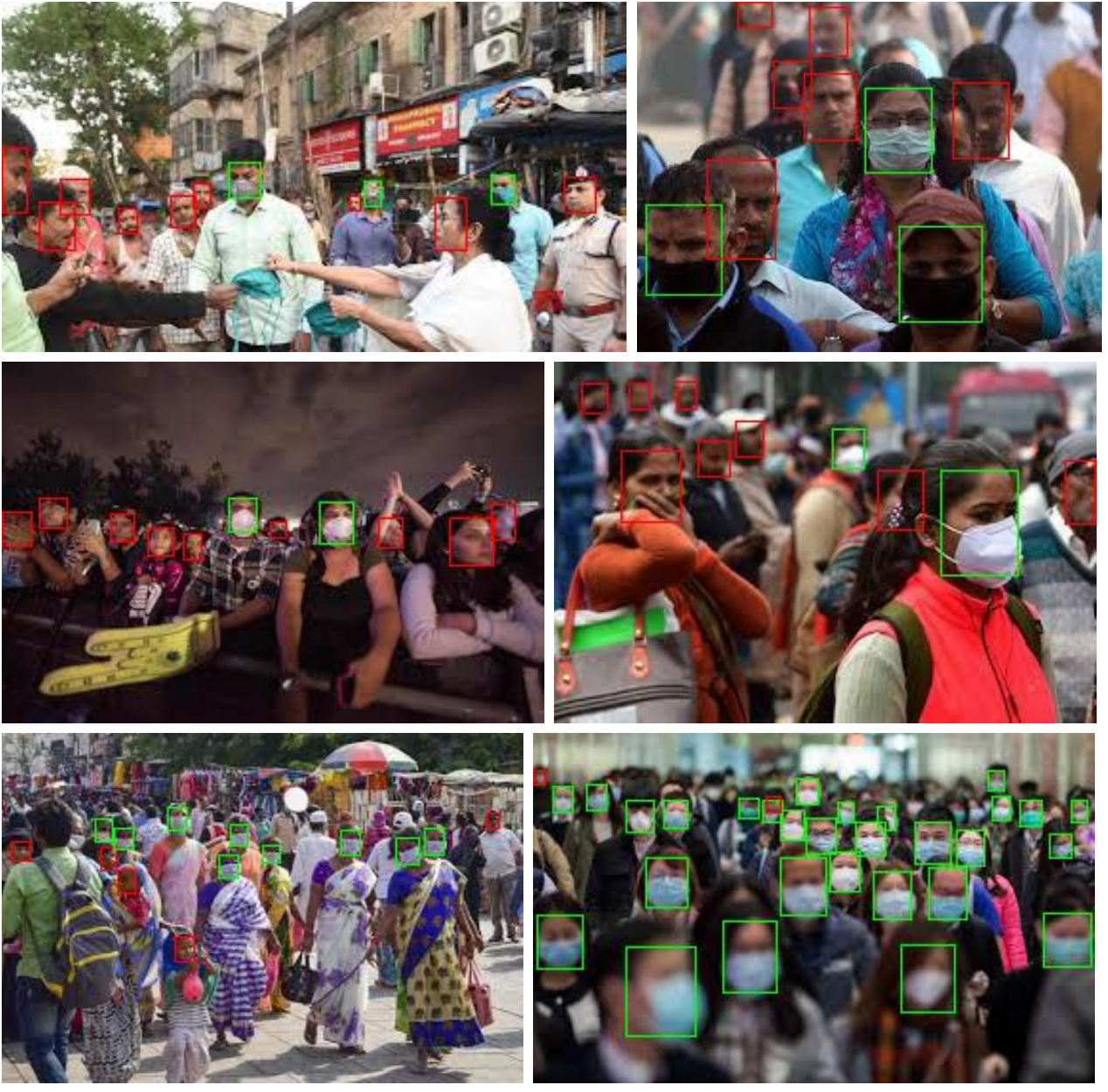}
    \caption{Sample images of the test set of the Moxa 3K dataset with predicted mask labels from the proposed pipeline where the green bounding box means face wearing a mask and the red bounding box means face without a mask.}
    \label{fig:output_images}
\end{figure}
\subsubsection{Results on Moxa 3K dataset} \label{subsec:moxa_3k_result}
Table \ref{tab:moxa_3k_result} compares the performance of the proposed pipeline with several state-of-the-art approaches, i.e., SSD, YOLOv3Tiny YOLOv3, F-RCNN proposed by Roy \textit{et al.} \cite{roy2020}, RetinaFaceMask by Jiang \textit{et al.} \cite{Jiang2020}, and SL-FMDet from Fan \textit{et al.} \cite{fan2021} on the Moxa 3K dataset.
This table shows that our pipeline achieves state-of-the-art performance on the Moxa 3K dataset by surpassing state-of-the-art approaches.
Compared to Roy \textit{et al.} \cite{roy2020} which simply fine-tuned popular objectors on this face mask dataset, our pipeline with lightweight detection backbone, MobileNetV2$_{1.0}$, can achieve higher mAP than F-RCNN which is havier and consumes higher computation resources.
Our pipeline yields a higher mAP than RetinaFaceMask about 0.68 points when using the same MobileNet backbone.
This result demonstrates the effectiveness of the two-step approach which can gain better results even having a smaller or similar backbone.
Moreover, our pipeline outperforms SL-FMDet on the Moxa 3K dataset by gaining 2.14 points higher on mAP.
Since Moxa 3K dataset is more challenging than the AIZOO dataset by having various scenarios from close-up faces to crowded scenes, our two-step approach which exploits face detection trained on a large-scale dataset is more robust than the single-step approaches.
Figure \ref{fig:output_images} shows sample images from the test set of the Moxa 3K dataset with predicted mask classes on the detected faces.
This figure demonstrates our pipeline's ability on the wide range of images taken in public areas.
\begin{table}[htbp]
\caption{The comparative result with the state-of-the-arts on the test set of the Moxa 3K dataset.}
\centering
\begin{tabular}{llc}
\toprule
\multicolumn{2}{l}{Method}                                                                                                       & mAP   \\
\midrule
\multicolumn{2}{l}{SSD \cite{roy2020} }                                                                                                          & 46.52 \\
\multicolumn{2}{l}{YOLOv3Tiny \cite{roy2020} }                                                                                                   & 56.27 \\
\multicolumn{2}{l}{YOLOv3 \cite{roy2020} }                                                                                                       & 66.84 \\
\multicolumn{2}{l}{F-RCNN \cite{roy2020} }                                                                                                       & 60.50 \\
\multicolumn{2}{l}{RetinaFaceMask - MobileNet \cite{Jiang2020} }                                                                                                       & 64.93 \\
\multicolumn{2}{l}{SL-FMDet \cite{fan2021}}                                                                                                     & 67.04 \\
\midrule
\multicolumn{3}{c}{\textbf{Results of our proposed pipeline}}                                                                            \\
\midrule
Face detector                                                  & \begin{tabular}[c]{@{}l@{}}Face mask classifier\end{tabular} & mAP   \\
\midrule
\begin{tabular}[c]{@{}l@{}}PyramidKey - MobileNetV2$_{0.25}$ \cite{earp2019}\end{tabular} & \multirow{4}{*}{MobileNetV1$_{1.0}$}                                    & 59.97 \\
\begin{tabular}[c]{@{}l@{}}PyramidKey - MobileNetV2$_{1.0}$ \cite{earp2019}\end{tabular} &                                                                 & 65.61 \\
\begin{tabular}[c]{@{}l@{}}RetinaFace - ResNet50 \cite{deng2019}\end{tabular}    &                                                                 & 68.79 \\
\begin{tabular}[c]{@{}l@{}}PyramidKey - ResNet152 \cite{earp2019}\end{tabular}   &                                                                 & \textbf{69.18} \\
\bottomrule
\end{tabular}
\par
\label{tab:moxa_3k_result}
\end{table}
\section{Conclusions}\label{sec:conclusion}
We propose a two-step face mask detection pipeline consisting of face detection with alignment and face mask classification modules. 
This approach allows us to exploit the state-of-the-art face detection model from RetinaFace \cite{deng2019} and PyramidKey \cite{earp2019}.
In the face mask classification module, we trained the model on our collected dataset, SertisFaceMask, which contains a wide variety of mask types with a balance of mask classes.

We provide the evidence to show that our proposed pipeline is superior to the state-of-the-art approaches of both single-step and two-step face mask detection on the AIZOO and Moxa 3K datasets.
Our pipeline yields an mAP of 93.80, which performs comparable with SL-FMDet and outperforms other state-of-art methods on the AIZOO dataset.
Moreover, we provide a relabeled annotation of the test set of the AIZOO dataset, and our pipeline can achieve state-of-the-art performance with the mAP of 94.58 on the relabeled test set.
Our pipeline also achieves state-of-the-art performance on Moxa 3K dataset with the highest mAP of 69.18, outperforming other compared methods.
The result on both AIZOO and Moxa 3K datasets shows the advantages of the two-step approach structure, efficient face detection, and face mask classification models.

\section*{Acknowledgment}

The authors express our gratitude to Aubin Samacoits and Christina Kim for their input and feedback during the writing of this paper.
The funding for the development of the pipeline and the use of the Sertis' face mask detection pipeline was
supported by the Health Systems Research Institute (HSRI).

\bibliographystyle{IEEEtran}
\bibliography{ms}

\end{document}